# Using Answer Set Programming in an Inference-Based approach to Natural Language Semantics

Farid Nouioua

LIPN UMR 7030 du C.N.R.S.
Institut Galilée – Univ. Paris-Nord
93430 Villetaneuse – FRANCE
nouiouaf@lipn.univ-paris13.fr

Pascal Nicolas

LERIA
University of Angers
2, bd Lavoisier F-49045 Angers cedex
pascal.nicolas@univ-angers.fr

## 1. Motivation

The traditional tri-partition syntax/semantics/pragmatics is commonly used in most of the computer systems that aim at the simulation of the human understanding of Natural Language (*NL*). This conception does not reflect the flexible and creative manner that humans use in reality to interpret texts. Generally speaking, formal *NL* semantics is referential i.e. it assumes that it is possible to create a static discourse universe and to equate the objects of this universe to the (static) meanings of words. The meaning of a sentence is then built from the meanings of the words in a compositional process and the semantic interpretation of a sentence is reduced to its logical interpretation based on the truth conditions. The very difficult task of adapting the meaning of a sentence to its context is often left to the pragmatic level, and this task requires to use a huge amount of common sense knowledge about the domain. This approach is seriously challenged (see for example [4][14]). It has been showed that the above tri-partition is very artificial because linguistic as well as extra-linguistic knowledge interact in the same global process to provide the necessary elements for understanding. Linguistic phenomena such as polysemy, plurals, metaphors and shifts in meaning create real difficulties to the referential approach of the *NL* semantics discussed above. As an alternative solution to these problems, [4] proposes an inferential approach to the *NL* semantics in which words trigger inferences depending on the context of their apparition. In the same spirit we claim that understanding a *NL* text is a reasoning process based on our knowledge about the norms[1] of its domain i.e. what we generally expect to happen in normal situations. But what kind of reasoning is needed for natural language semantics?

The answer to this question is based on the remark that texts seldom provide normal details that are assumed to be known to the reader. Instead, they focus on abnormal situations or at least on events that cannot be inferred by default from the text by an ordinary reader. A central issue in the human understanding of *NL* is the ability to infer systematically and easily an amount of implicit information necessary to answer indirect questions about the text. The consequences resulting from truth-based entailments are logically valid but they are poor and quite limited. Those obtained by a norm-based approach are defeasible: they are admitted as long as the text does not mention explicit elements that contradict them. However they provide richer information and enable a deeper understanding of the text. That is why the norm-based reasoning must be non-monotonic. In addition to this central question, the representation language must take into account a number of modalities (including the temporal aspect) that are very useful to answer different questions on *NL* texts.

The next section gives a general logical framework to represent in a first order language the necessary knowledge about a domain and allows non-monotonic reasoning. Section 3 shows how to implement our representation language fragment in the formalism of Answer Set Programming by transforming them into extended logic programs. In section 4, we discuss the use of our language in the car crash domain to find automatically the cause of an accident from its textual description. The

1 In A.I, the word norm is commonly used in the « normative » sens. Here, it is rather used in the « normal » sens.

kind of inference rules required in this application is showed through a detailed presentation of the analysis of a text from the corpus we are using. Finally, we conclude and give some perspectives for future work in section 5.

## 2. Knowledge representation language

The explicit information evoked in a given text provides the starting point for the reasoning process that aims to understand it. Thus, the first task to do is to extract from the text this explicit information and to represent it in an adequate language. The richness and flexibility of *NL* constrains the representation language to take into account a number of aspects whose necessity and importance may vary from an application to another. In what follows, we describe a logical language which enhances within the first order framework some aspects that we believe to be useful in an inferential approach to *NL* semantics. Namely, the proposed language allows the representation of time, modalities and non-monotonic inferences (see [7] for more details).

### 2.1 Reification

The first idea that comes to mind when representing knowledge about *NL* statements is to use first order predicates to express properties of objects, agents …etc. However we need often to treat further aspects. For example, we need to represent modalities on the considered properties and to reason about them i.e. to use the predicate names themselves as variables over which one can quantify in order to avoid the use of ad hoc inference rules, i.e. to factorise the rules at an adequate level of abstraction. To solve this problem within the framework of first order logic, we use the reification technique, commonly used in Artificial Intelligence (*AI*). Instead of writing *P(X, Y)* to express the fact that property *P* applies to arguments *X* and *Y*, we write *Holds(P, X, Y)*. The property name *P* becomes then an argument in the new predicate *Holds*. i.e. P will be a variable over properties and it can be quantified in inference rules.
The use of the reification technique yields to two main drawbacks: first, it forces a fixed arity for the predicate Holds whereas properties in general may have a different number of arguments. The second problem is the necessity to redefine ad hoc axioms about the properties (negation, conjunction, disjunction… of properties). One possible solution to the first problem is to consider a special binary function *combine* which constructs a new "complex" argument from two other arguments. For example, as the predicate *Holds* has three arguments then, the predicate *Q(X, Y, Z)* can be reified as : *Holds(combine(Q, X), Y, Z)*[2]. In general, this corresponds well to linguistic practice: for example the application of a transitive verb to its complement can be considered as a unique "complex" property comparable to an intransitive verb. Concerning the second problem, it turns out that in practice we often do not need all the axioms but only some particular ones. So we have to represent only those axioms that we really need in the application considered.

### 2.2 Representing time

Generally, narrative texts describe events that take place in a time perceived as continuous. The temporal aspect is crucial in their understanding. Two representation approaches are possible for time: either we represent the continuous time which reflects the physical reality and use the elegant mathematical tools developed for mechanics, or we represent the discrete time which reflects the text structure and which corresponds rather to a naive physics. We chose the second approach, because generally, texts are written by persons who ignore the mathematical details of motion, and they can be understood without having such knowledge. Two approaches are still possible for a discrete model of time. Either we use a linear model in which only the events that happened in reality are represented, or we take into account the unrealized futures as part of the temporal

2 As a concrete example, the ternary predicate bump(A, B, T) (vehicle A bumps vehicle B at time T) is written after reification and by using the combine function as : Holds(combine(bump, B), A, T). The term combine(bump, B) expresses then the complex property of « bumping the vehicle B ».

information. In this case, we use a branching time model [5][10]. This last model is richer than the former and can be very useful in some cases. In this paper we are interested only on the linear model. What is important for us in time modelling is to establish an order between the events evoked in the text. Of course, this choice limits the use of our language to applications which do not need deeper structure of time but it remains useful in practice (see section 4 for a possible application). Indeed, the unrealized futures are not completely excluded in our model, as they can be represented implicitly by modalities (see the modality *able* in section 4.2.2 ).
The semantics used for time in our model is situated somehow between an interval-based and a point-based semantics: the scene of the accident described in the text is decomposed as a succession of ordered time elements. Each time element is denoted by an integer representing its order number. This integer is used as an argument in the predicates. The meaning of the element depends on the nature of the property. If it is a persistent property, the time parameter denotes the entire time interval during which this property remains true (interval based semantics). If the property is not persistent (corresponds to an action or a punctual event) then the temporal argument denotes the starting point of the interval on which the property occurs and causes at least one persistent property to change its truth value.

### 2.3 Modalities

Modalities express properties of the predicates other than their truth value, which can be considered as a null modality. Different types of modal logics have been developed to formalize the reasoning about modalities. Even though the reasoning we want to apply on texts makes use of modalities, it can be carried out without developing new modal logics with 'complete' axiomatizations. What we really need is to represent the modalities as first order predicates using the reification technique discussed in section 2.1., and to define only useful axioms as inference rules. For example, to represent the fact that the modality *Mod* is applied to the predicate *P* having $X_1, \ldots, X_n$ as arguments we write : $Mod(P, X_1, \ldots, X_n)$ instead of the classical notation : $Mod\ P(X_1, \ldots, X_n)$.

### 2.4 Non-monotonicity

Non-monotonicity is an essential characteristic of the nature of the reasoning used by humans to understand texts. Among the different approaches proposed in the literature to formalise this variant of commonsense reasoning, we have used Reiter's default logic [11] to represent our inference rules. The fixed point semantics used to compute the default theories extensions seems to be adequate to the nature of the *NL* understanding process. Indeed, as discussed in section 1, the *NL* understanding process cannot be decomposed in a sequence of separate steps but it consists in the simultaneous satisfaction of several linguistic as well extra-linguistic constraints in a manner that can be approached by the search of some fixed point of the meaning of the given text.
Two kinds of inference rules are considered: the strict inferences represented by material implications and the defeasible ones represented by Reiter's defaults. To facilitate the implementation of our rules on the answer set programming paradigm (see section 3) we limit their forms as follows:

Let $A_1, \ldots, A_n, B, C_1, \ldots, C_k$ be first order literals.
The Expression (1) is a material implication. It means that *B* is inferred whenever $A_1, \ldots, A_n$ are verified. Two kinds of default rules are considered. The first form (2) corresponds to a "normal" default. It means that if we have $A_1, \ldots, A_n$ then, we can infer *B* as long as this is consistent. The second one (3) corresponds to a semi-normal default and its meaning is that in general, when we have $A_1, \ldots, A_n$ then, we can infer *B* as long as this is consistent and none of $\neg C_i$ $(i=1..k)$ belongs to the extension[3]. Semi normal defaults are particularly useful to establish a priority order between inference rules which can not be done using only normal defaults[12].

---

3 We use a notation in which A : B stands for $\frac{A : B}{B}$ and A : B[C] stands for $\frac{A : B, C}{B}$

$$A_1 \wedge ... \wedge A_n \rightarrow B \quad (1)$$
$$A_1 \wedge ... \wedge A_n : B \quad (2)$$
$$A_1 \wedge ... \wedge A_n : B[C_1, ..., C_k] \quad (3)$$

## 3. Implementation by Answer Set Programming

### 3.1. Theoretical backgrounds

Answer Set Programming *(ASP)* is a recent paradigm covering different kinds of logic programs, and associated semantics. It allows representing and solving various problems in Artificial Intelligence. On one hand, we can cite combinatorial problems as k-coloring graph, path finding, timetabling, ... On another hand, *ASP* is also concerned by problems arising when available information is incomplete as non-monotonic reasoning, planning, diagnosis, ... The non familiar reader will find additional information about *ASP* on the web site of the working group *WASP* (http://wasp.unime.it/).

In the present work we are particularly interested in using *ASP* as a framework for default reasoning. For this we use Extended Logic Programs *(ELP)* to represent knowledge by means of rules containing positive information and strong or default negative information and we interpret them by answer set semantics [3]. Formally, an *ELP* is a set of rules of the form

$$c \leftarrow a_1, ..., a_n, \text{not } b_1, ..., \text{not } b_m. \qquad n \geq 0 \text{ and } m \geq 0$$

where $c$, ai and $b_j$ are literals.
For a given rule $r$, we denote

$$head(r) = c \quad body^+(r)=\{a_1, ..., a_n\} \quad body^-(r)=\{b_1, ..., b_m\} \quad r^+=c \leftarrow a_1, ..., a_n$$

**Definition** Let $R$ be a set of rules without default negation $(\forall r \in R, body^-(r) = \emptyset)$, $R$ is called a Definite Logic Program. A literal set $X$ is closed wrt $R$ when $\forall r \in R, body^+(r) \subseteq X \Rightarrow head(r) \in X$. The set of consequences of $R$ is $C_n(R)$ the minimal literal set that is closed wrt $R$ consistent or equal to the whole set of literals of the language

For a given literal set $A$ and an *ELP* $P$, the reduct of $P$ by $A$ is the definite Logic Program

$$P^A=\{r^+ \mid r \in P \text{ and } body^-(r) \cap A = \emptyset\}$$

**Definition** Let $P$ be an *ELP* and $A$ a literal set. $A$ is an answer set of $P$ if and only if $A=Cn(P^A)$

**Examples**

$P1=\{a \leftarrow \text{not } b., b \leftarrow \text{not } a., \neg c \leftarrow b.\}$ has two answer sets $\{a\}$ and $\{b, \neg c\}$
$P2=\{a \leftarrow \text{not } a.\}$ has no answer set at all.

We have recalled the basic notions of answer set semantics only in the case of propositional rules. But, obviously, for a more flexible knowledge representation, rules may contain variables. In this case, a rule is considered as a global schema for the set of fully instanciated rules that can be obtained by replacing every variable by every constant in the language.

**Example**

$P=\{bird(1)., bird(2)., penguin(2)., fly(X) \leftarrow bird(X), \text{not } penguin(X)., \neg fly(X) \leftarrow penguin(X).\}$ is equivalent to the program $P'=\{bird(1)., bird(2)., penguin(2)., fly(1) \leftarrow bird(1), \text{not } penguin(1)., \neg fly(1) \leftarrow penguin(1)., fly(2) \leftarrow bird(2), \text{not } penguin(2)., \neg fly(2) :\leftarrow penguin(2).\}$
Then, $P$ (formally $P'$) has one answer set $\{bird(1), bird(2), penguin(2), fly(1), \neg fly(2)\}$.
Let us mention an important point for our work that is answer set semantics for *ELP* can be viewed as a subcase of default logic [2][3]. By translating every rule $r = c \leftarrow a_1, ..., a_n, \text{not } b_1, ..., \text{not } b_m.$ into the default rule : $T(r) = a_1 \wedge ... \wedge a_n : c\ [\neg b_1, ..., \neg b_m]$

By this way :

If *S* is an answer set of an *ELP P*, then *Th(S)* is an extension of the default theory *(∅,T(P))* every extension of *(∅,T(P))* is the deductive closure of one answer set of *P*.

Obviously, in whole generality every default theory cannot be translated into an *ELP*. But as we explain it later, it is possible to encode some restricted default theories in an *ELP*. By this way it is possible to envisage realistic applications of default reasoning since several software packages for *ASP* are available today, e.g. the following ones:

*DLV*[8] http://www.dbai.tuwien.ac.at/proj/dlv,
*Smodels* [13] http://www.tcs.hut.fi/Software/smodels
*Cmodels* [9] http://www.cs.utexas.edu/users/tag/cmodels.html
*Nomore*++[1] http://www.cs.uni-potsdam.de/wv/nomore++

### 3.2. From Default Logic to ASP

Here, we explain how we have encoded our knowledge base that is originally a default theory, into an extended logic program. A very important point to note is that our original knowledge base does not contain disjunctions. Since a default theory is a pair consisting in a set of classical formulas and a set of default rules, we distinguish two major translations.

| ***classical formulas*** | ***ELP*** |
|---|---|
| one fact : *a* | one rule with an empty body : *a*. |
| a conjunction of n facts : $a_1 \wedge ... \wedge a_n$ | n rules with empty bodies: $a_1$. ... $a_n$. |
| a material implication $a_1 \wedge ... \wedge a_n \rightarrow b$ | one direct rule $b \leftarrow a_1, ... , a_n$.<br>and n contrapositive rules :<br>$\neg a_1 \leftarrow \neg b, a_2, ... , a_n$.<br>...<br>$\neg a_n \leftarrow \neg b, a_1, ... , a_{n-1}$. |

| ***default rules*** | ***ELP*** |
|---|---|
| $A_1, ..., A_n : B$<br>$A_1, ..., A_n : B[C_1, ..., C_k]$ | $b \leftarrow a_1, ..., a_n, not\ \neg b$.<br>$b \leftarrow a_1, ..., a_n, not\ \neg b, not\ \neg c_1,..., not\ \neg c_m$. |

We have preferred to encode firstly our rules in default logic instead using directly *ASP* because default logic is more compact than *ASP,* which needs more rules, especially for contrapositives. The translation of default logic into *ASP* can be easily auomated.

## 4. From the description of an accident to its cause

### 4.1. The corpus

We are working on a sample of 60 representative texts of a larger corpus. These texts are short descriptions of car accident circumstances. They are written (in French) by persons implied in the accidents to be sent to their insurance company[4]. The length of our texts varies between 9 and 167 words. They contain 129 sentences whose length varies between 4 and 55 words; the longest report has 7 sentences and there are 24 reports that contain only one sentence. The total number of word occurrences is 2256. But there are only 500 distinct words corresponding to 391 dictionary entries.

4 We are grateful to the MAIF insurance company for having given us access to the reports that constitute our corpus.

### 4.2. Our task

#### 4.2.1. Finding the cause of the accident

The objective of the system we are developing is to find automatically the cause of an accident from its textual description. Because of the very controversial nature of causality we must define more precisely our objective. We are interested in our study by the interventionist conception of causality in which voluntary actions are privileged as potential causes of events. This is in correspondence with the practical use of causality in *AI.* Moreover, we claim that the most plausible causes for abnormal situations like accidents are those that reflect violation of norms (anomalies)[6]. We consider that the system has understood a text if it finds the same cause as the one given by an ordinary human reader. We have then determined manually the cause of each text and we have used this information to validate the results of the system.

Two essential steps are considered in the overall architecture of the system. The first one "the linguistic step" applies a tagger and syntactical analyser to extract a set of surface relations between words. These relations are then progressively transformed by an adequate " linguistic reasoning" into the so-called "semantic predicates" which express the explicit information provided by the text. The semantic predicates are represented in a "semantic language" as the one discussed in section 2. This part of the system, which is under construction, tries to adapt existing methods to deal with the problems of anaphora resolution and time ordering of the events described in a text. We will not discuss the details of the linguistic step in this paper. The second step: "the semantic step" applies a set of strict and default inference rules based on norms of the road domain to enrich the semantic predicates initially extracted from the text by further semantic predicates enhancing implicit information. The inference rules are designed manually and reflect rudimentary reasoning that any reader of the text makes systematically. This semantic reasoning process stops as soon as the system infers the necessary information that characterizes an anomaly. Section 5 gives further details about the semantic reasoning through an example taken from the corpus.

#### 4.2.2. Some specificities

The majority of the semantic predicates used in our system have the form: *Holds(P, A, T)* where *P* is a simple or a complex property (expressed by the binary function combine), *A* is an agent (generally a vehicle involved in the accident) and *T* is the order number of a time interval during which (or at the beginning of which) the property *P* holds (to simplify, we will say henceforth that property *P* holds at time *T*). For example *Holds(stop, ag, 3)* means that the agent '*ag*' is stopped at time 3 and $Holds(combine(follows, ag_1), ag_2, 2)$ means that at time 2, agent '$ag_2$' follows agent '$ag_1$' (in a file of vehicles). When needed a function *neg* is applied to a property to have its negation. We introduce the rule (4)

$$Holds(neg(P), A, T) \leftrightarrow \neg Holds(P, A, T) \qquad (4)$$

The main modalities that we use in our system cope respectively with duties and capacities :

*must(P, A, T)* means that at time *T*, agent *A* has the duty to achieve the property *P*.
*able(P, A, T)* means that at time *T*, agent *A* is able to achieve the property *P*. In terms of branching time, this means that there is some possible future in which *P* holds.

The semantic reasoning is designed so that it converges to a "kernel" containing a limited number of semantic predicates[5] in terms of which all possible anomalies can be expressed. In a given text, it is possible that several anomalies coexist. In this case, the system distinguishes between the primary anomaly which can be considered as the most plausible cause of the accident and the other anomalies called "derived anomalies". A primary anomaly has two forms: either an agent *A* has the duty and the capacity to achieve a property *P* at a time *T* and at time *T+1* a property *P′* incompatible

5 The predicates of the kernel are : *Holds(control, A, T)* [*A* has the control of his/her vehicle], *Holds(moves_back, A, T)* [*A* moves back], *Holds(starts, A, T)* [*A* moves off], *Holds(drives_slowly, A, T)* [ *A* drives fairly slowly], *Holds(stops, A, T)* [*A* is stopped], *Holds(comb(disruptive_factor, X), A, T)* [*X* is a disruptive factor for *A*] with *P* holds (5) or some disruptive and inevitable factor occurs and causes the accident (6). The form of a derived anomaly (7) differs from that of a primary one only on the agent's capacity.

*primary_an(P, A, T) ← property(P), vehicle(A), time(T), must(P, A, T), able(P, A, T), holds(P', A, T+1), incompatible(P, P')* (5)

*primary_an(combine(disruptive_factor, X), A, T) ← object(X), vehicle(A), time(T), holds(combine(disruptive_factor, X), A, T)* (6)

*derived_an(P, A, T) ← property(P), vehicle(A), time(T), must(P, A, T), ¬ able(P, A, T), holds(P', A, T+1), incompatible(P, P')* (7)

### 4.3. An example

To illustrate our methodology, let us consider the following text of the corpus (translated into english) and explain the inference rules involved in its analysis :

*« Whereas vehicle B was overtaking me, the driver lost the control of its vehicle. It bumped on the central guardrail , and crossed the ways. It then cut my way. My vehicle A initially bumped on vehicle B on its right side, before being crushed on the guardrail. »*

The set of the semantic predicates extracted from the text are :

*holds(overtake, veh_b, 1), ¬ holds(control, veh_b, 2),*
*holds(combine(bump, guardrail), veh_b, 3), ¬ holds(stop, veh_b, 4),*
*holds(combine(bump, veh_b),veh_a, 5), holds(combine(bump, guardrail), veh_a, 6)*
*vehicle(veh_a), vehicle(veh_b), object(veh_a), object(veh_b), object(guardrail).*

In what follows, we show how the application of inference rules leads to the determination of the primary and the derived anomalies:

Rule(8) states that *"at the starting state 0, each vehicle has the control"*.

*holds(control, A, 0) ← agent(A), vehicle(A)* (8)

It allows to infer : *holds(control, veh_a, 0), holds(control, veh_b, 0)*

Rule(9) states that *"if B is a vehicle that bumps on A at time T, then B is not stopped at this time"*.

*¬ holds(stop, A, T) ← vehicle(A), object(B), time(T), holds(combine(bump, B), A, T)* (9)

It allows to infer: *¬ holds(stop, veh_b, 3), ¬ holds(stop, veh_a, 5), ¬ holds(stop, veh_a, 6)*

Rules(10) and (11) state that *"if A is a vehicle that bumps on B at time T, then there is at this time a shock (symmetric) between A and B"*.

*holds(combine(shock, B), A, T) ← vehicle(A), object(B), time(T), holds(combine(bump, B), A, T)* (10)

*holds(combine(shock, A), B, T) ← object(A), object(B), time(T), holds(combine(shock, B), A, T)* (11)

The set of predicates inferred by these rules are :

*holds(combine(shock, guardrail), veh_b, 3), holds(combine(shock, veh_b), guardrail, 3),*
*holds(combine(shock, veh_b), veh_a, T), holds(combine(shock, veh_a), veh_b, T),*
*holds(combine(shock, veh_a), guardrail, T), holds(combine(shock, guardrail), veh_a, T)*

Rule(12) states that *"if A is implied in two successive shocks at times T and T+1, then we deduce that it lost the control after the first shock (during the time interval T)"*.

*¬ holds(control, A, T) ← agent(A), object(B), object(C), time(T), holds(combine(shock, A), B, T), holds(combine(shock, A), C, T+1)* (12)

It allows to infer: *¬ holds(control, veh_a, 5)*

The remainder of information about the control of vehicles *A* and *B* during the other time intervals are deduced using appropriate rules that handle the persistence of some particular properties. The complete set of conclusions concerning control is as follows :

*holds(control, veh_b, T) (for 0≤ T ≤ 1), ¬ holds(control, veh_b, T) (for 2≤ T ≤ 6), holds(control, veh_a, T) (for 0≤ T ≤ 4), ¬ holds(control, veh_a, T) (for 5≤ T ≤ 6)*

Rule(13) states that *"in general if there is a collision between a vehicle A and an object B at time T, then B represents an obstacle for A at time T-1".*

*holds(combine(obstacle, A), B, T-1) ← object(A), vehicle(B), time(T), holds(combine(shock, A), B, T), not ¬ holds(combine(obstacle, A), B, T-1)* *(13)*

We obtain from this rule :

*holds(combine(obstacle, guardrail), veh_b, 1), holds(combine(obstacle, veh_a), veh_b, 4), holds(combine(obstacle, veh_b), veh_a, 4), holds(combine(obstacle, guardrail), veh_a, 5)*

Rules (14) and (15) allows to infer that some obstacles are not predictable. The rule (14) states that *"if a vehicle B not controlled represents at time T an obstacle to vehicle A, then this obstacle is not predictable for A at this time T".* Whereas rule (15) states that *"in general, if a vehicle B bumps a vehicle A at time T, then B is considered as an umpredictable obstacle for A at time T".*

*¬ predictable(combine(obstacle, B), A, T) ← vehicle(B), vehicle(A), time(T), holds(combine(obstacle, B), A, T), ¬ holds(control, B, T)* *(14)*

*¬ predictable(combine(obstacle, B), A, T) ← vehicle(A), vehicle(B), instant(T), vrai(combine(bump, A), B, T), not predictable(combine(obstacle, B), A, T)* *(15)*

By these two rules we can infer : *¬ predictable(combine(obstacle, veh_a),veh_b, 4), ¬ predictable(combine(obstacle, veh_b), veh_a, 4)*

Rule(16) states that *"in general, one must keep the control of one's vehicle "*

*must(control,A,T) ← vehicle(A), time(T), not ¬ must(control,A,T), not ¬ holds(control,A,T)* *(16)*

This rule infers : *must(control, veh_b, 1), must(control, veh_a, 4)*

The meaning of rule(17) is that *"one must avoid any obstacle".*

*must(combine(avoid, X), A, T) ← vehicle(A), object(X), time(T), holds(combine(obstacle, X), A, T)* *(17)*

This rule infers : *must(combine(avoid, guardrail), veh_b, 1), must(combine(avoid, veh_a), veh_b, 4) must(combine(avoid, veh_b), veh_a, 4), must(combine(avoid, guardrail), veh_a, 5)*

Rule(18) states that *"in general the duty to avoid an obstacle turns out to the duty to stop (this default is inhibited by a number of situations illustrated in the rule)"*

*must(stop, A, T) ← vehicle(A), object(B), time(T), must(combine(avoid, B), A, T), holds(combine(shock, B), A, T+1), not ¬ must(stop, A, T), not must(drive_slowly, A, T), not holds(stop, A, T), not holds(combine(follow, A), B, T), not must(not(backwards), A, T-1), not must(not(move_off), A, T-1), not ¬ predictable(combine(obstacle, B), A, T)* *(18)*

We can infer from this rule : *must(stop, veh_b, 1), must(stop, veh_a, 5)*
Rules (19) and (20) are the main rules that allow to infer agent's capacities :

*able(P, A, T) ← vehicle(A), object(B), time(T), action(Act), property(P), pcb(Act, P), available(Act, P, A, T)* *(19)*

*¬able(P, A, T) ← vehicle(A), object(B), time(T), action(Act), property(P), pcb(Act, P), ¬available(Act, P, A, T)* *(20)*

they mean that "vehicle A is able to reach property P at time Tn if and only if there is some action Act which is a "potential cause" for P and which is available for A to reach P at time T (the *contrapositives ar*e omitted)".

The occurrences of the relation *pcb* (which abreviates: potentially caused by) are statically determined and stored in a static database. In our case we have : *pcb(brake, stop), pcb(combine(keep_state, control)*[6]*, control).*

By default, actions are available for agents to reach the corresponding properties. This default inference is inhibited by a number of strict rules. In our case, we obtain *:*

*available(combine(keep_state, control), control, veh_b, 1) (the default is applied)*
*¬available(combine(keep_state, control), control, veh_a, 4)*[7]
*¬available(brake, stop, veh_a, 5)*[8]

From these results it follows :

*able(control, veh_b, 1), ¬ able(stop, veh_a, 4), ¬ able(stop, veh_a, 5).*

The application of rules (5) and (7) we can detect the primary and the derived anomalies :
*primary_an(control, veh_b, 1), derived_an(control, veh_a, 4), derived_an(stop, veh_a, 5)*

Finally, the cause of the accident is expressed by: "the loss of control of vehicle B at time 1"

## 5. Conclusion and perspectives

This paper defends the idea that inferences are at the heart of the problematic of NL semantics. We have showed that the inferences we need to understand natural language are based on our knowledge about the norms of the domain and are non-monotonic since the conclusions of this kind of reasoning are in general defeasible. We proposed a general representation language which takes into account within a first order framework modalities, time and non-monotonicity that are essential aspects in an inferential approach of NL understanding. We presented also how to transform our inference rules into extended logic programs. To illustrate our approach in a practical domain we have used a corpus of 60 short texts describing the circumstances of road accidents. We have used *Smodels* to implement our reasoning system. With about 200 inference rules, the system succeeds to find for each text only one stable model containing the necessary literals which express the primary and the derived anomalies. We have determined manually for each text the answer that we hope to obtain. Thus, the validation criterion is that the system gives for each text the same answer as the predetermined one. The running time varies from a text to another but it does not exceed 30 seconds which is rather encouraging. Many other perspectives of future work are open, among them:

- Analyzing more texts of the same domain in order to verify :
    - The validity of our hypotheses, especially those concerning the relationship between norms and causes and the sufficiency of a linear model of time;
    - that the inference rules have a sufficient degree of generality to be adapted easily to new situations by giving the expected answers for new reports.
    - the adequacy of the proposed representation language to deal with new texts.
- Generalizing the approach to other domains

6 we consider as action the fact of keeping holded some persistent property.
7 the lost of control because of a shock at time T makes unavailable the action of keeping the control at time T-1.
8 if a vehicle is not under control, then, any action is unavailable for its driver.

**Acknowledgment.** The authors are indebted to Daniel Kayser for very helpful remarks on previous versions of this text.